\documentclass[11pt]{article}
\PassOptionsToPackage{hyperfootnotes=false}{hyperref}
\usepackage[preprint]{acl}
\usepackage{times}
\usepackage{latexsym}
\usepackage[T1]{fontenc}
\usepackage[utf8]{inputenc}
\usepackage{microtype}
\usepackage{graphicx}
\usepackage{amsmath,amssymb}
\usepackage{booktabs}

\hypersetup{
  pdftitle={Fine-Grained Units, Flat Decisions: Hierarchical Micro-Expert Pruning for Mixture-of-Experts Models},
  pdfauthor={Zukang Xu, Xinpei Gao, Xing Hu, Zhe Jiang, Dawei Yang},
  pdfsubject={UC-RAMP: post-training micro-expert pruning for mixture-of-experts models},
  pdfkeywords={Mixture-of-Experts, Structured Pruning, Model Compression, Expert Capacity Allocation, Channel Selection}
}

\title{Fine-Grained Units, Flat Decisions: Hierarchical Micro-Expert Pruning for Mixture-of-Experts Models}
\author{
  \normalsize Zukang Xu\textsuperscript{1}\thanks{Equal contribution.}\quad
  Xinpei Gao\textsuperscript{1,2}\footnotemark[1]\quad
  Xing Hu\textsuperscript{1,3}\\[3pt]
  \normalsize\bfseries Zhe Jiang\textsuperscript{3}\quad
  Dawei Yang\textsuperscript{1}\thanks{Corresponding author: \href{mailto:dawei.yang@houmo.ai}{\nolinkurl{dawei.yang@houmo.ai}}.}\\[5pt]
  {\normalfont\small \textsuperscript{1}Houmo AI}\\
  {\normalfont\small \textsuperscript{2}University of Science and Technology of China}\\
  {\normalfont\small \textsuperscript{3}Southeast University}
}
\date{}

\begin{document}
\maketitle
\begin{abstract}
Micro-expert pruning reduces the capacity of mixture-of-experts (MoE) models by removing coupled channels within routed experts. However, finer pruning granularity alone does not guarantee the preservation of downstream capabilities, as similar language-modeling perplexity can accompany substantially different task performance. We revisit micro-expert mask construction as two distinct decisions: allocating retained capacity across experts and selecting channels within that capacity. We propose \textbf{UC-RAMP}, a two-stage post-training pruning framework that explicitly separates these decisions. CAPE uses cross-domain calibration evidence to redistribute expert widths under an exact layer-wise budget, while RAMP refines channel composition at fixed widths through one-out, one-in exchanges guided by a channel-interaction surrogate. Across three MoE models at 25\% and 50\% routed-expert channel pruning, UC-RAMP achieves the highest ten-task mean accuracy among evaluated pruning methods in five of six settings. At 50\% pruning, it outperforms the strongest evaluated baseline by 8.19, 3.96, and 6.14 percentage points across the three models, respectively. In a single-A100 deployment at 25\% pruning, UC-RAMP reduces measured time to first token by 19.7--23.9\% and time per output token by 20.2--22.3\% relative to the unpruned model. The code is available at \url{https://github.com/xzktx003/UC_RAMP}.

\end{abstract}

\section{Introduction}

Sparse mixture-of-experts (MoE) models reduce per-token computation by routing each input to only a subset of experts while retaining a substantially larger parameter pool \citep{shazeer2017outrageously,fedus2022switch}. However, sparse activation does not eliminate the storage and deployment costs of the full expert pool. Post-training pruning can reduce these costs without retraining the model from scratch. Existing pruning methods operate at different granularities, ranging from individual weights \citep{frantar2023sparsegpt,sun2024wanda} and coupled model dimensions \citep{ma2023llmpruner,ashkboos2024slicegpt} to entire experts \citep{lu2024notall,lasby2026reap}. While whole-expert pruning removes all channels of a selected expert, micro-expert pruning instead removes corresponding gate and up-projection rows together with a down-projection column \citep{xu2026camera}, allowing expert capacity to be reduced without necessarily removing the entire expert.

\begin{figure*}[t]
\centering
\includegraphics[width=\textwidth]{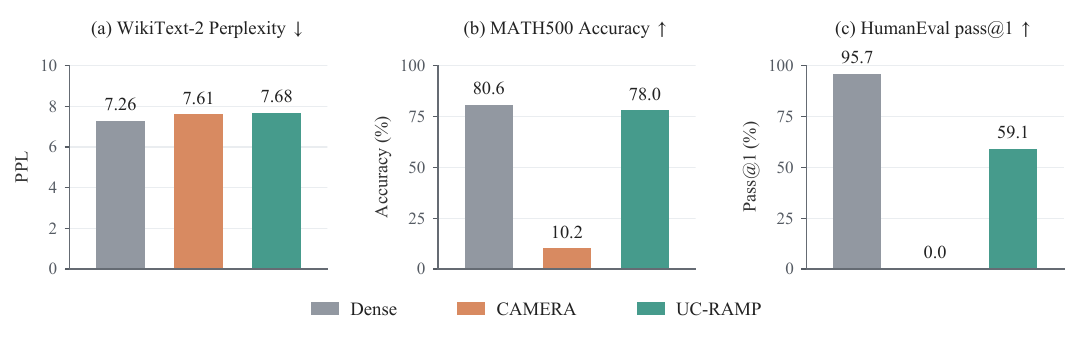}
\caption{Perplexity and downstream task performance at 50\% routed-expert channel pruning on Qwen3-30B-A3B-Instruct-2507. CAMERA exhibits substantial degradation on MATH500 and HumanEval despite a small change in WikiText-2 perplexity. At the same nominal pruning ratio, UC-RAMP achieves substantially higher task accuracy. Lower perplexity and higher task accuracy indicate better performance.}
\label{fig:motivation}
\end{figure*}

Channel-level pruning offers greater flexibility in constructing structural masks, but finer granularity alone does not guarantee the preservation of downstream capabilities. Figure~\ref{fig:motivation} illustrates this limitation on Qwen3-30B-A3B-Instruct-2507. At 50\% routed-expert intermediate-channel pruning, CAMERA's WikiText-2 perplexity remains close to that of the dense model (7.61 versus 7.26), whereas MATH500 accuracy drops from 80.6\% to 10.2\% and HumanEval pass@1 from 95.7\% to 0\%. In contrast, UC-RAMP achieves substantially higher task accuracy under the same pruning budget, while exhibiting similarly low perplexity. These observations highlight the importance of mask construction beyond pruning granularity alone, motivating a closer examination of how individual channel scores are translated into a structural pruning mask.

A micro-expert pruning mask encodes two distinct structural decisions: \textbf{capacity allocation}, which determines the number of channels retained in each expert, and \textbf{channel composition}, which specifies the identities of those channels. Under a fixed layer-wise channel budget, a global Top-$K$ ranking determines both decisions simultaneously: selecting a channel not only includes it in the retained set but also allocates a portion of the budget to its expert. However, these decisions can benefit from different sources of information. Cross-domain response statistics can guide expert-capacity allocation, whereas interactions among channels provide complementary information for evaluating channel sets at fixed widths. While prior methods such as MoE-Slimming consider nonuniform expert widths \citep{ding2026attribution}, our focus is on explicitly separating capacity allocation from channel composition, allowing each decision to be guided by its corresponding evidence.

To operationalize this separation, we introduce UC-RAMP, a two-stage micro-expert pruning framework consisting of Capability-Aware Pruning via Envelopes (CAPE) and Reconstruction-Aware Micro-expert Pruning (RAMP). In the first stage, CAPE establishes a shared channel ranking from cross-domain calibration statistics and evaluates each expert on a common reference channel set. An expert-level multiplier then adjusts the competition for retained capacity, enabling width redistribution under an exact layer-wise budget while preserving the channel ranking within each expert. In the second stage, RAMP freezes the allocated widths and incorporates pairwise channel interactions into a reconstruction-based surrogate. It then refines the retained channel sets through one-out, one-in exchanges that reduce the surrogate without changing expert widths. Both stages operate without gradient updates or modifications to the router.

Our main contributions are summarized as follows:
\begin{itemize}
\setlength{\itemsep}{2pt}
\setlength{\parsep}{0pt}
\item \textbf{We formulate micro-expert pruning as two distinct structural decisions:} allocating retained capacity across experts and selecting channels within the allocated capacity. This formulation makes explicit the two decisions jointly determined by conventional global channel ranking.
\item \textbf{We propose UC-RAMP, a two-stage post-training pruning framework.} CAPE uses cross-domain calibration evidence to redistribute expert capacity under an exact channel budget, while RAMP incorporates channel interactions to refine retained channel sets without changing expert widths.
\item \textbf{We evaluate UC-RAMP across three MoE models and two pruning ratios.} UC-RAMP achieves the highest mean accuracy among the evaluated pruning methods in five of six settings, including all three models at 50\% pruning. Component ablations characterize the contributions of the two stages, and deployment measurements demonstrate inference latency reductions with nonuniform expert widths.
\end{itemize}

\section{Related Work}

\paragraph{Post-training and whole-expert pruning.}
Post-training pruning has been widely studied for reducing the storage and computational costs of large language models. SparseGPT \citep{frantar2023sparsegpt} and Wanda \citep{sun2024wanda} perform weight-level pruning using reconstruction information and weight-activation saliency, respectively, while LLM-Pruner \citep{ma2023llmpruner} and SliceGPT \citep{ashkboos2024slicegpt} remove coupled model dimensions. For MoE models, another line of work focuses on expert-level compression. \citet{lu2024notall} investigate expert pruning and skipping, REAP \citep{lasby2026reap} evaluates routed expert contributions, and DiEP \citep{bai2025diep} optimizes nonuniform expert pruning across layers. More recently, AIMER \citep{liu2026aimer} employs calibration-free weight statistics, whereas MAN/MSAN \citep{liu2026score} leverages routed-token activation statistics. These approaches primarily operate at the granularity of complete experts, in contrast to micro-expert pruning, which allows capacity to be reduced within individual experts.

\paragraph{Expert-internal pruning and dependent channel selection.}
Finer-grained MoE compression retains greater flexibility in selecting expert-internal structures. MoE-Pruner \citep{xie2024moepruner} incorporates routing information into expert-internal parameter pruning, while CAMERA \citep{xu2026camera} introduces micro-experts as coupled channels across the gate, up, and down projections of SwiGLU experts. MoE-Slimming \citep{ding2026attribution} further considers nonuniform expert widths through attribution-guided and coverage-based structural compression. Beyond independent importance estimates, The Sparsity Whisperer \citep{kong2026sparsity} highlights the importance of dependencies among jointly pruned units. Building on this line of research, UC-RAMP explicitly separates expert-capacity allocation from fixed-width channel composition. CAPE uses cross-domain responses evaluated on shared reference channel sets to redistribute expert widths under an exact layer-wise budget, whereas RAMP incorporates pairwise channel interactions to refine channel membership without changing the allocated widths.

\section{Method}
\label{sec:method}

\subsection{Capacity and Composition of Micro-Expert Masks}

Consider a routed expert $e$ in MoE layer $\ell$ with intermediate width $C$. For a token representation $x_t$, the activation of its $c$-th SwiGLU channel is
\begin{equation}
z_{t,e,c} =
\operatorname{SiLU}(w_{g,e,c}^{\top}x_t)
(w_{u,e,c}^{\top}x_t).
\end{equation}
Here and below, we omit the layer index in channel activations and weight vectors for clarity. Let $r_{t,e}$ denote the routing weight used in the model's forward pass. The routed expert contribution can be expressed as
\begin{equation}
r_{t,e}E_e(x_t)
= \sum_{c=1}^{C} r_{t,e}z_{t,e,c}w_{d,e,c}.
\end{equation}
The corresponding gate and up-projection rows and down-projection column constitute a micro-expert, which is removed as a single structured unit.

Let $m_{\ell,e,c}\in\{0,1\}$ indicate whether channel $c$ is retained. The retained width of expert $e$ and the exact layer-wise channel budget are defined as
\begin{equation}
k_{\ell,e}=\sum_{c=1}^{C}m_{\ell,e,c},
\qquad
\sum_e k_{\ell,e}=K_\ell.
\label{eq:budget}
\end{equation}
Here $K_\ell$ is the retained channel budget in layer $\ell$. A pruning mask therefore specifies two structural decisions: \textbf{capacity allocation}, represented by the expert widths $k_{\ell,e}$, and \textbf{channel composition}, represented by the retained sets $S_{\ell,e}=\{c:m_{\ell,e,c}=1\}$, where $|S_{\ell,e}|=k_{\ell,e}$.

\begin{figure*}[t]
\centering
\includegraphics[width=0.94\textwidth]{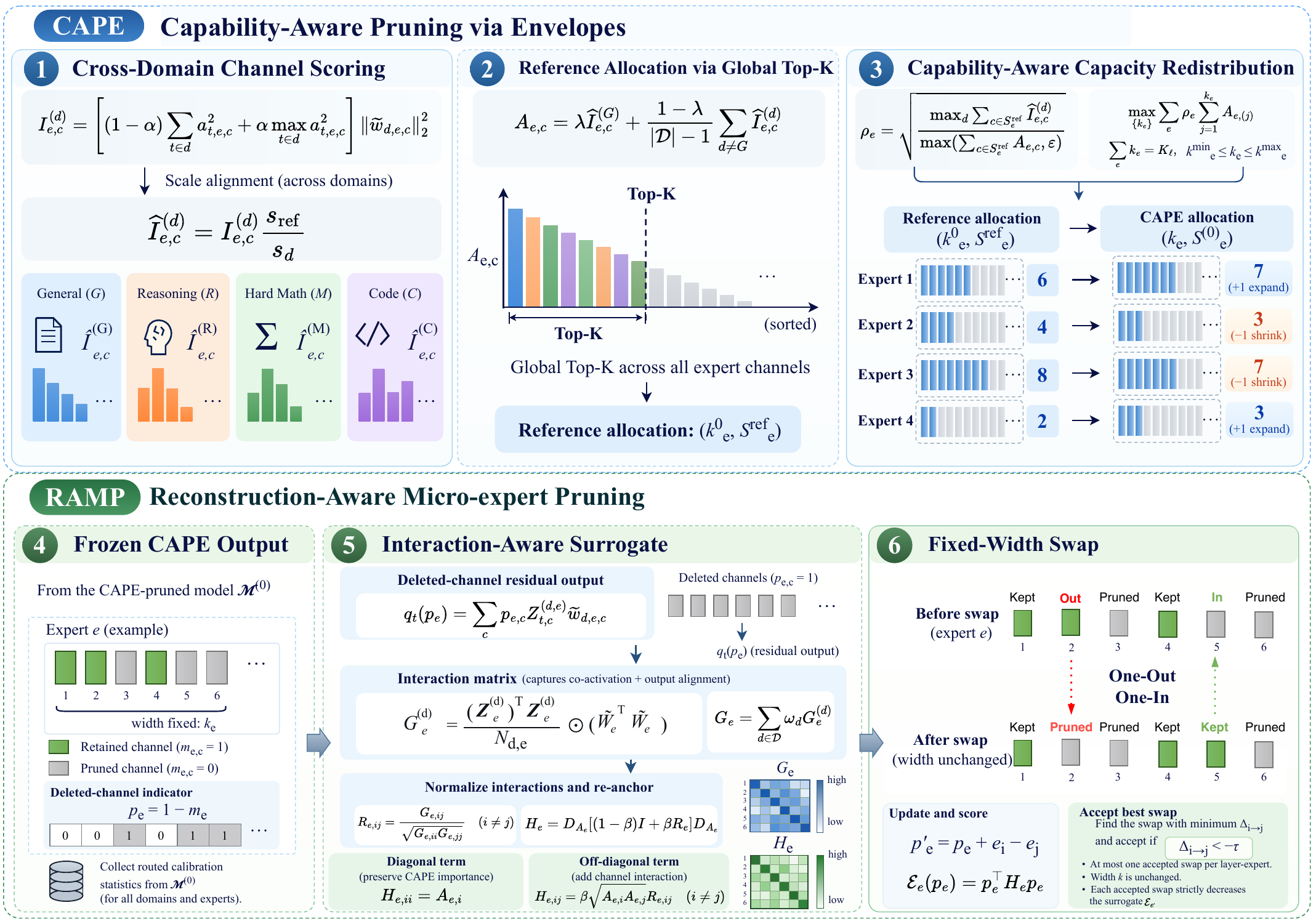}
\caption{Overview of UC-RAMP. CAPE combines cross-domain channel importance with expert-level capacity adjustment to allocate retained widths under an exact layer-wise budget. RAMP then constructs an interaction-aware surrogate and refines channel composition through one-out, one-in exchanges while preserving the allocated widths.}
\label{fig:method}
\end{figure*}

A global Top-$K_\ell$ ranking of individual channel scores determines both decisions simultaneously. Changing an expert's scores can redistribute capacity across experts even when the channel ordering within that expert remains unchanged. Conversely, exchanging a retained channel with a pruned channel within the same expert changes its composition without affecting its width. UC-RAMP explicitly separates these operations: CAPE determines the expert widths and initial retained sets, whereas RAMP refines channel composition while preserving the allocated widths. Figure~\ref{fig:method} illustrates this two-stage interface.

\subsection{CAPE: Allocating Expert Capacity}

CAPE redistributes retained channels across experts under a fixed layer-wise budget $K_\ell$. It first constructs a cross-domain channel ranking and then adjusts expert-level capacity competition without changing the channel ordering within each expert.

\paragraph{Cross-domain channel importance.}
Let $\mathcal{D}$ denote the calibration domains, including a general-language domain $G$ and complementary domains. For tokens routed to expert $e$, we define $a_{t,e,c}=\bar r_{t,e}z_{t,e,c}$, where $\bar r_{t,e}$ is the routing probability normalized over the selected experts. To account for the scale of each channel's output contribution, we weight its down-projection vector by the next layer's RMSNorm coefficient $\gamma^{(\ell+1)}$, obtaining $\widetilde w_{d,e,c}=\gamma^{(\ell+1)}\odot w_{d,e,c}$. The domain-specific channel importance combines accumulated response energy and peak response:
\begin{equation}
\begin{split}
I_{e,c}^{(d)}={}&
\left[(1-\alpha)\sum_{t\in d}a_{t,e,c}^{2}\right.\\
&\left.+\alpha\max_{t\in d}a_{t,e,c}^{2}\right]
\left\|\widetilde w_{d,e,c}\right\|_2^2.
\end{split}
\label{eq:cape_importance}
\end{equation}

Because calibration domains differ in content and routing patterns, their raw importance scores may have different scales. Within each layer, we compute $s_d=\operatorname{mean}_{e,c}I_{e,c}^{(d)}$ and the geometric reference $s_{\mathrm{ref}}=(\prod_{d\in\mathcal{D}}s_d)^{1/|\mathcal{D}|}$. We then normalize each domain as $\widehat I_{e,c}^{(d)}=I_{e,c}^{(d)}s_{\mathrm{ref}}/s_d$. This alignment equalizes domain-wise mean scores while preserving their relative channel rankings.

The normalized scores are combined into a general-language-weighted channel anchor:
\begin{equation}
A_{e,c}
=
\lambda\widehat I_{e,c}^{(G)}
+
\frac{1-\lambda}{|\mathcal{D}|-1}
\sum_{d\ne G}\widehat I_{e,c}^{(d)}.
\label{eq:cape_anchor}
\end{equation}

The anchor $A_{e,c}$ provides a unified ranking of individual channels across calibration domains.

\paragraph{Expert-level capacity adjustment.}
Within layer $\ell$, a global Top-$K_\ell$ selection based on $A$ defines reference widths $k_e^0$ and retained channel sets $S_e^{\mathrm{ref}}$, satisfying $\sum_e k_e^0=K_\ell$. We evaluate every calibration domain on the same expert-specific reference set and define the expert multiplier
\begin{equation}
\rho_e=\sqrt{\frac{\max_d\sum_{c\in S_e^{\mathrm{ref}}}\widehat{I}_{e,c}^{(d)}}{\max(\sum_{c\in S_e^{\mathrm{ref}}}A_{e,c},\varepsilon)}}.
\label{eq:cape_multiplier}
\end{equation}

For an expert with zero reference width, its highest-ranked anchor channel is used only to compute the multiplier, without changing the reference allocation.

The maximum cross-domain response forms a capability envelope over the expert's reference channels. The multiplier $\rho_e$ compares this envelope with the fused anchor response on the same channel set, providing an expert-level signal for capacity redistribution. Applying it uniformly to an expert's channels gives $T_{e,c}=\rho_e A_{e,c}$. For $\rho_e>0$, this reweighting changes competition for capacity across experts while preserving the channel ranking within each expert. Thus, cross-domain evidence adjusts expert-level capacity priorities without redefining the channel anchor.

\paragraph{Budget-constrained width allocation.}
Let $A_{e,(j)}$ denote the $j$-th largest anchor score within expert $e$. To constrain capacity migration from the reference allocation, we impose lower and upper width bounds $k_e^{\min}$ and $k_e^{\max}$ around $k_e^0$. CAPE determines the retained widths by solving
\begin{equation}
\begin{aligned}
\max_{\{k_e\}}\quad&
\sum_e \rho_e \sum_{j=1}^{k_e} A_{e,(j)}
\\
\mathrm{s.t.}\quad&
\sum_e k_e=K_\ell,\qquad k_e\in\mathbb{Z},
\\
&
k_e^{\min}\le k_e\le k_e^{\max}.
\end{aligned}
\label{eq:cape_allocation}
\end{equation}

The bounds define a local trust region around the reference widths. Since the weighted marginal scores are nonincreasing within each expert, the objective is optimized exactly by retaining all lower-bound prefixes and selecting the largest eligible remaining marginal scores until the budget is filled.

CAPE outputs the allocated widths $k_e$ and initial retained sets $S_e^{(0)}=\operatorname{TopK}(A_e,k_e)$. The width bounds and the optimality argument are detailed in Appendices~\ref{app:protocol} and~\ref{app:proofs}, respectively. These widths remain fixed during the subsequent RAMP refinement.

\subsection{RAMP: Fixed-Width Channel Refinement}

CAPE constructs initial retained sets using individual channel importance scores, without explicitly accounting for interactions among simultaneously removed channels. RAMP addresses this limitation by refining channel composition while keeping the expert widths determined by CAPE fixed. To estimate these interactions, RAMP collects routed calibration statistics from the intermediate model $\mathcal M^{(0)}$ obtained after applying CAPE's initial pruning masks.

Let $m_e\in\{0,1\}^{C}$ denote the retained-channel mask of expert $e$, with the layer index omitted, and let $p_e=\mathbf{1}-m_e$ indicate deleted channels. For tokens from domain $d$ routed to expert $e$, define $Z_{t,c}^{(d,e)}=r_{t,e}z_{t,e,c}$, where $r_{t,e}$ is the routing weight used in the forward pass. Under fixed channel responses, the output contribution of the deleted channels is
\begin{equation}
q_t(p_e)=
\sum_c p_{e,c}Z_{t,c}^{(d,e)}
\widetilde w_{d,e,c}.
\label{eq:ramp_residual}
\end{equation}
Its mean squared norm can be written as $p_e^\top G_e^{(d)}p_e$, where
\begin{equation}
G_e^{(d)}=\frac{(Z_e^{(d)})^\top Z_e^{(d)}}{N_{d,e}}\odot(\widetilde W_e^\top\widetilde W_e).
\label{eq:ramp_gram}
\end{equation}

Here $N_{d,e}$ is the number of routed calibration tokens, and $\widetilde W_e$ contains the scaled down-projection vectors as columns. We aggregate the domain-specific Gram matrices as
\begin{equation}
G_e=\sum_{d\in\mathcal D}\omega_d G_e^{(d)},
\label{eq:ramp_aggregation}
\end{equation}
where $\omega_G=\lambda$ and $\omega_d=(1-\lambda)/(|\mathcal D|-1)$ for $d\ne G$, following the same domain weighting scheme as CAPE in Eq.~\eqref{eq:cape_anchor}.
The off-diagonal entries of $G_e$ capture channel co-activation and the alignment of their down-projection outputs.

Rather than directly using the diagonal energies of $G_e$, RAMP preserves CAPE's individual channel importance while incorporating the interaction structure of $G_e$.

We obtain a normalized Gram matrix $R_e$ by rescaling the pairwise interaction entries of $G_e$:
\begin{equation}
R_{e,ij}=
\begin{cases}
1, & i=j,\\[2pt]
\frac{G_{e,ij}}{\sqrt{G_{e,ii}G_{e,jj}}},
& \begin{gathered}
i\ne j,\\
G_{e,ii}G_{e,jj}>0,
\end{gathered}\\[4pt]
0, & \text{otherwise}.
\end{cases}
\label{eq:ramp_correlation}
\end{equation}
Let $D_{A_e}=\operatorname{diag}(\sqrt{A_e})$. We then construct the interaction-aware surrogate:
\begin{equation}
\begin{aligned}
H_e
&=
D_{A_e}\bigl[(1-\beta)I+\beta R_e\bigr]D_{A_e},
\\[2pt]
\mathcal{E}_e(p_e)
&=
p_e^\top H_e p_e.
\end{aligned}
\label{eq:ramp_surrogate}
\end{equation}

By construction, the diagonal and off-diagonal entries of $H_e$ satisfy
\begin{equation}
\begin{aligned}
H_{e,ii}&=A_{e,i},\\[2pt]
H_{e,ij}&=\beta\sqrt{A_{e,i}A_{e,j}}R_{e,ij}
\quad (i\ne j).
\end{aligned}
\label{eq:ramp_entries}
\end{equation}
The diagonal preserves CAPE's individual channel importance, whereas the off-diagonal terms incorporate pairwise channel interactions, controlled by $\beta$. The resulting quadratic form evaluates deleted-channel sets using both sources of information.

RAMP minimizes this surrogate locally through fixed-width channel exchanges. For a retained channel $i$ and a deleted channel $j$, the candidate update is $p'_e=p_e+\mathbf{e}_i-\mathbf{e}_j$. We evaluate the exact surrogate change $\Delta_{i\to j}=\mathcal{E}_e(p'_e)-\mathcal{E}_e(p_e)$ and accept the best candidate only if $\Delta_{i\to j}<-\tau$, where $\tau\geq 0$. The main configuration permits at most one accepted exchange per layer-expert. Each exchange preserves $k_e$, and the acceptance criterion guarantees a strict decrease in the defined surrogate. For $0\leq\beta\leq 1$, $H_e$ is positive semidefinite, as shown in Appendix~\ref{app:proofs}.

The two stages follow a fixed execution order. CAPE applies the initial pruning masks sequentially from shallow to deep layers, producing the intermediate model $\mathcal M^{(0)}$. RAMP then collects calibration statistics using this model and refines channel composition without changing the expert widths allocated by CAPE. The exact exchange formula and execution details are provided in Appendices~\ref{app:proofs} and~\ref{app:protocol}, respectively.

\section{Experiments}
\label{sec:experiments}

\subsection{Models, Budgets, and Evaluation}

We evaluate UC-RAMP on Qwen3-30B-A3B-Instruct-2507, Gemma-4-26B-A4B-it, and Qwen3.6-35B-A3B at pruning ratios of 25\% and 50\%. These ratios refer exclusively to the intermediate channels of routed experts, excluding attention, embeddings, routers, shared experts, Gemma's parallel dense MLP, and multimodal modules. We distinguish channel-level and expert-level pruning baselines in Table~\ref{tab:main}, as identical nominal pruning ratios can lead to different structural configurations and runtime characteristics.

UC-RAMP uses four calibration domains, each consisting of sequences of length 2048: 128 WikiText-2 training examples for general language \citep{merity2017pointer}; 64 GSM8K and 64 MathQA training examples for general reasoning \citep{cobbe2021gsm8k,amini2019mathqa}; 128 competition-MATH training examples for advanced mathematics \citep{hendrycks2021math}; and 64 MBPP training examples for code \citep{austin2021mbpp}. Calibration requires only forward-pass statistics, without gradient updates. We set the general-domain weight to $\lambda=0.9$, distributing the remaining weight equally among the three specialized domains. The other hyperparameters are fixed at $\alpha=0.9$ in Eq.~\eqref{eq:cape_importance} and $\beta=0.2$ in Eq.~\eqref{eq:ramp_surrogate}. Details of the width trust-region constraints are provided in Appendix~\ref{app:protocol}.

Our channel-level baselines include CAMERA \citep{xu2026camera} and MoE-Slimming \citep{ding2026attribution}, while the whole-expert baselines comprise AIMER \citep{liu2026aimer}, REAP \citep{lasby2026reap}, and MAN/MSAN \citep{liu2026score}. We also report results for the unpruned models. The baseline results retain the respective scoring procedures and calibration settings of each method. Table~\ref{tab:ablation} further disentangles the contributions of CAPE's capacity redistribution and RAMP's fixed-width channel refinement.

All evaluations are conducted using EvalScope 1.4.1 with temperature set to 0, a maximum generation length of 2048 tokens, and a fixed random seed of 1234. We report WikiText-2 perplexity separately from accuracy on ten downstream benchmarks: PIQA \citep{bisk2020piqa}, ARC-Easy and ARC-Challenge \citep{clark2018arc}, GSM8K \citep{cobbe2021gsm8k}, HellaSwag \citep{zellers2019hellaswag}, WinoGrande \citep{sakaguchi2020winogrande}, RACE \citep{lai2017race}, GPQA-Diamond \citep{rein2024gpqa}, MATH500 \citep{hendrycks2021math,lightman2024verify}, and HumanEval pass@1 \citep{chen2021evaluating}. The average accuracy reported in Tables~\ref{tab:main} and~\ref{tab:ablation} is the unweighted mean across the ten benchmarks, excluding perplexity. Detailed per-task results are provided in Appendix~\ref{app:fullresults}.

\subsection{Results across models and budgets}

\begin{table*}[t]
\centering
\small
\begin{tabular}{llrrrrrr}
\toprule
& & \multicolumn{2}{c}{Qwen3-30B-A3B} & \multicolumn{2}{c}{Gemma-4-26B} & \multicolumn{2}{c}{Qwen3.6-35B-A3B}\\
\cmidrule(lr){3-4}\cmidrule(lr){5-6}\cmidrule(lr){7-8}
Method & Unit & 25\% & 50\% & 25\% & 50\% & 25\% & 50\%\\
\midrule
Dense & none & 86.19 & 86.19 & 90.15 & 90.15 & 76.94 & 76.94\\
CAMERA & channel & 68.63 & 58.46 & 69.12 & 56.12 & 69.90 & 56.13\\
MoE-Slimming & channel & 78.43 & 67.00 & 79.85 & 63.92 & 64.71 & 49.62\\
AIMER & expert & 80.82 & 64.46 & 62.27 & 41.83 & 70.97 & 48.89\\
REAP & expert & 79.29 & 66.30 & 79.74 & 27.92 & 69.37 & 65.41\\
MAN & expert & 81.89 & 64.82 & 69.96 & 49.21 & 74.26 & 68.09\\
MSAN & expert & 79.94 & 67.24 & 83.99 & 48.08 & \textbf{75.59} & 67.05\\
\midrule
UC-RAMP & channel & \textbf{82.81} & \textbf{75.43} & \textbf{84.96} & \textbf{67.88} & 67.18 & \textbf{74.23}\\
\bottomrule
\end{tabular}
\caption{Ten-task mean accuracy (\%) for UC-RAMP and pruning baselines. Columns specify nominal pruning ratios, and Unit identifies each method's pruning granularity. UC-RAMP's ratios apply to routed-expert intermediate channels. Bold denotes the highest pruned-model mean in each setting.}
\label{tab:main}
\end{table*}

\paragraph{Overall Performance.}
UC-RAMP achieves the highest ten-task mean accuracy among pruned models in five of the six settings in Table~\ref{tab:main}, including all three models at 50\% pruning. At this ratio, it reaches 75.43\% on Qwen3-30B-A3B, 67.88\% on Gemma-4-26B, and 74.23\% on Qwen3.6-35B-A3B. These scores exceed the respective strongest baselines, MSAN (67.24\%), MoE-Slimming (63.92\%), and MAN (68.09\%), by 8.19, 3.96, and 6.14 percentage points.

\paragraph{Capability Recovery.}
On Qwen3 at 50\% pruning, UC-RAMP achieves near-dense MATH500 accuracy: 78.0\%, compared with 80.6\% for the unpruned model and 10.2\% for CAMERA. HumanEval pass@1 shows substantial recovery over CAMERA (59.15\% versus 0\%), compared with 95.73\% for the unpruned model. WikiText-2 perplexities remain similar: 7.68 for UC-RAMP and 7.61 for CAMERA, compared with 7.26 for the unpruned model. These task-level gains at similar perplexity address the capability loss highlighted in Figure~\ref{fig:motivation}.

\paragraph{Performance Across Architectures.}
The advantages at 50\% pruning extend to Gemma-4 and Qwen3.6, although recovery varies across individual tasks and architectures. Their MATH500 accuracies reach 73.8\% and 52.8\%, compared with CAMERA's 4.4\% and 12.2\%. HumanEval pass@1 reaches 14.02\% and 75.61\%, compared with zero for CAMERA on both models. Qwen3.6 at 25\% pruning is the single exception in Table~\ref{tab:main}: UC-RAMP obtains 67.18\% mean accuracy, while MSAN leads with 75.59\%. Appendix~\ref{app:fullresults} provides the complete task scores.

\subsection{Ablation Studies}
\label{sec:ablations}

\subsubsection{Component Ablation}
\label{sec:component_ablation}

Table~\ref{tab:ablation} isolates expert-capacity redistribution and fixed-width channel refinement while retaining multi-domain scoring. At 50\% pruning, adding CAPE's expert multipliers and capacity redistribution raises mean accuracy from 73.11\% to 75.19\% on Qwen3 and from 65.51\% to 67.60\% on Gemma, gains of 2.08 and 2.09 percentage points. RAMP then improves the means to 75.43\% and 67.88\% with these expert widths held fixed. Appendix~\ref{app:ablations} reports the other four CAPE--UC-RAMP comparisons.

\begin{table}[t]
\centering
\small
\begin{tabular}{lccr}
\toprule
Model & CAPE & RAMP & Mean ACC (\%)\\
\midrule
Qwen3 & $\times$ & $\times$ & 73.11\\
 & $\checkmark$ & $\times$ & 75.19\\
 & $\checkmark$ & $\checkmark$ & \textbf{75.43}\\
\midrule
Gemma-4 & $\times$ & $\times$ & 65.51\\
 & $\checkmark$ & $\times$ & 67.60\\
 & $\checkmark$ & $\checkmark$ & \textbf{67.88}\\
\bottomrule
\end{tabular}
\caption{Component ablation at 50\% routed-expert channel pruning. $\times/\times$ denotes Multi-domain Top-$K$, retaining UC-RAMP's multi-domain scoring and calibration; it is distinct from CAMERA. The CAPE column toggles expert multipliers and capacity redistribution, while RAMP toggles fixed-width swaps. Bold marks the highest mean for each model.}
\label{tab:ablation}
\end{table}

These results distinguish the contributions of the two structural decisions. The multi-domain Top-$K$ baseline already provides a channel ranking, while CAPE additionally changes how the retained-channel budget is distributed across experts. RAMP operates within these allocated widths and provides a smaller incremental gain through interaction-aware set refinement. The larger improvement from capacity redistribution in both settings suggests that expert width allocation is an important factor in constructing effective micro-expert masks.

\subsubsection{Hyperparameter Sensitivity}
\label{sec:hyperparameters}

\begin{figure*}[t]
\centering
\includegraphics[width=\textwidth]{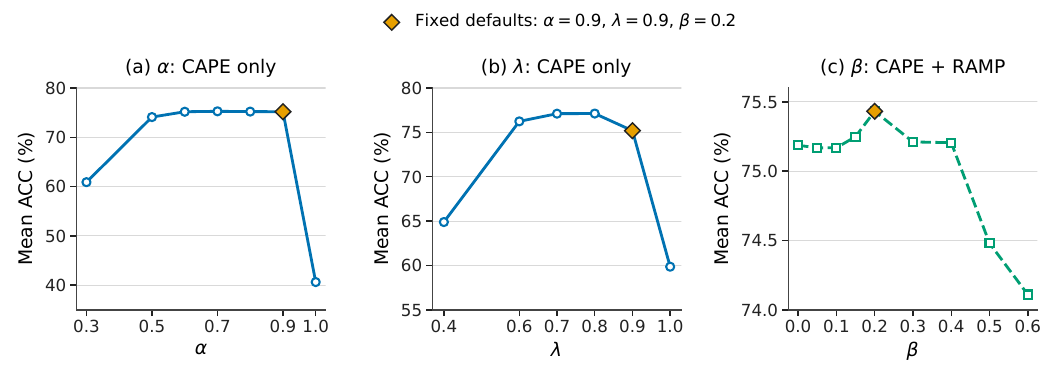}
\caption{Hyperparameter sensitivity on Qwen3-30B-A3B-Instruct-2507 at 50\% routed-expert channel pruning. Panels (a) $\alpha$ and (b) $\lambda$ report CAPE-only results. Panel (c) $\beta$ reports RAMP initialized from CAPE, with expert widths fixed. Diamonds mark the fixed defaults $(0.9,0.9,0.2)$. The panels use separate y-axis ranges.}
\label{fig:hyperparameters}
\end{figure*}

We examine the three hyperparameters on Qwen3-30B-A3B-Instruct-2507 at 50\% pruning (Figure~\ref{fig:hyperparameters}). The $\alpha$ and $\lambda$ sweeps evaluate CAPE only, while the $\beta$ sweep adds RAMP to the CAPE checkpoint. The main experiments use the fixed defaults $\alpha=0.9$, $\lambda=0.9$, and $\beta=0.2$.

Balancing activation energy and peak response yields similar mean accuracy across $\alpha\in[0.6,0.9]$, ranging from 75.178\% to 75.268\%. The tested endpoints, $\alpha=0.3$ and $\alpha=1.0$, give 60.891\% and 40.627\%, respectively, supporting a blend of the two response statistics.

The fusion weight $\lambda$ favors a balance between general and specialized evidence. The highest recorded CAPE mean is 77.119\% at $\lambda=0.8$, compared with 75.187\% at the fixed default $\lambda=0.9$. Both tested endpoints ($\lambda=0.4$ and $1.0$) score below 65\%.

For fixed-width refinement, moderate interaction strength improves mean accuracy from 75.187\% at $\beta=0$ to 75.432\% at $\beta=0.2$. Increasing the weight to $\beta=0.6$ yields 74.11\%. This trend supports incorporating channel interactions with a moderate weight in this setting.

\subsection{Inference Efficiency}
\label{sec:efficiency}

We compare unpruned and 25\%-pruned Qwen3 in BF16 using vLLM 0.27.1 (TP=1) on a single NVIDIA A100-SXM4 (80\,GB). UC-RAMP packs expert weights at their retained widths $k_e$, preserving the logical budget $\sum_e k_e=K_\ell$. Its custom Triton kernels use masked channel tiles (16 for gate/up and 32 for down reduction), while temporary activations are allocated by the maximum expert width in each layer. Table~\ref{tab:efficiency} includes these execution overheads.

Requests generate 32 tokens with temperature 0 and EOS ignored. Time to first token (TTFT) and time per output token (TPOT) are measured through local HTTP/SSE streaming using actual output token IDs. We report medians over two rounds, each with one warmup and three timed requests per configuration, reversing model startup order between rounds. TTFT decreases by 19.7--23.9\% and TPOT by 20.2--22.3\% relative to the dense reference. At the same 25\% pruning ratio, UC-RAMP retains 82.81\% ten-task mean accuracy, compared with 86.19\% for the unpruned model (Table~\ref{tab:main}). These measurements demonstrate lower inference latency under the evaluated deployment configurations, including custom-kernel and workspace overheads. Appendix~\ref{app:protocol} gives the deployment details.

\begin{table}[t]
\centering
\small
\setlength{\tabcolsep}{3pt}
\begin{tabular}{@{}rrrrrr@{}}
\toprule
 & & \multicolumn{2}{c}{TTFT (ms)} & \multicolumn{2}{c}{TPOT (ms/token)}\\
\cmidrule(lr){3-4}\cmidrule(lr){5-6}
Input & Batch & Dense & UC-RAMP & Dense & UC-RAMP\\
\midrule
256 & 1 & 106.61 & 81.71 & 47.45 & 36.94\\
512 & 1 & 107.92 & 82.14 & 47.54 & 36.92\\
1024 & 1 & 103.84 & 83.42 & 46.68 & 36.88\\
1024 & 2 & 131.38 & 102.79 & 48.48 & 38.70\\
1024 & 4 & 159.12 & 126.92 & 48.44 & 38.57\\
\bottomrule
\end{tabular}
\caption{Inference latency on Qwen3-30B-A3B-Instruct-2507 at 25\% routed-expert channel pruning, using one NVIDIA A100-SXM4 (80\,GB). Both models use BF16. Input is prompt length in tokens; Batch is batch size.}
\label{tab:efficiency}
\end{table}

\section{Conclusion}

Micro-expert pruning involves two structural decisions: allocating retained capacity across experts and selecting channels within that capacity. UC-RAMP separates these decisions through cross-domain capacity redistribution in CAPE and fixed-width channel refinement in RAMP. Across three MoE models and two pruning ratios, UC-RAMP achieves the highest mean accuracy among evaluated pruning methods in five of six settings, including all three models at 50\% pruning. Deployment measurements further demonstrate reduced inference latency at 25\% pruning. These results highlight the value of treating expert capacity and channel composition as distinct decisions in post-training MoE compression.

\section*{Limitations}

The experiments cover three MoE checkpoints and two pruning rates defined only over routed-expert intermediate channels. The calibration evidence spans four selected domains, so the observed allocation behavior may depend on which capabilities those domains represent. RAMP optimizes a local output-reconstruction surrogate within each expert; its guarantee concerns that surrogate, and its mixed task-level effects show that it should not be read as a general monotonic improvement to downstream accuracy.

\bibliography{references}
\clearpage
\appendix
\raggedbottom
\section{Implementation and protocol details}
\label{app:protocol}

The general-language calibration set contains 128 WikiText-2 training windows. The general-reasoning set contains 64 GSM8K and 64 MathQA training windows. The harder-mathematics set contains 128 sampled competition-MATH training windows, and the code set contains 64 MBPP training windows. Each window has length 2048. Each domain maintains separate channel statistics. The general domain receives weight 0.9 and the other three domains share the remaining 0.1 equally.

\begin{samepage}
For layer $\ell$, the reference ranking first assigns $k_{\ell,e}^{0}$ retained channels to expert $e$. With $C$ intermediate channels per expert and pruning rate $r_\ell$, the trust-region radius is
\begin{equation}
\Delta_\ell =
\max\!\left(1,\left\lfloor
\operatorname{round}[C(1-r_\ell)]/8
\right\rfloor\right).
\end{equation}
\end{samepage}
The allocation bounds are $k_{\ell,e}^{\min}=\max(0,k_{\ell,e}^{0}-\Delta_\ell)$ and $k_{\ell,e}^{\max}=\min(C,k_{\ell,e}^{0}+\Delta_\ell)$. CAPE processes layers from shallow to deep while retaining the masks already chosen for earlier layers. Once all CAPE masks are fixed, RAMP gathers routed trajectories from that checkpoint, computes within-expert Gram matrices, and considers at most one accepted exchange per layer-expert. Final masks are checked against every layer's exact budget.

\paragraph{Inference deployment.}
Table~\ref{tab:efficiency} compares the unpruned Qwen3-30B-A3B-Instruct-2507 checkpoint with UC-RAMP at 25\% routed-expert intermediate-channel pruning. Both models use BF16 without quantization on one NVIDIA A100-SXM4 (80\,GB), with tensor parallelism set to 1 in vLLM 0.27.1. Both use FlashAttention 2 and \texttt{enforce\_eager}, with \texttt{torch.compile} and CUDA Graph disabled. The maximum context length is 2048 tokens, and \texttt{gpu\_memory\_utilization} is 0.90.

The dense model uses vLLM's built-in \texttt{TritonExperts} with its default MoE configuration. The pruned model uses custom Triton kernels in two stages: gate/up projections with SwiGLU, followed by down projection and routing-weight aggregation. Expert weights are packed contiguously at their exact retained widths and located through offsets. The gate/up kernels use channel tiles of 16; the down kernel uses reduction tiles of 32 channels and output tiles of 64 hidden dimensions. Partial tiles are masked. Temporary activations and the gate/up launch grid are sized by the largest retained expert width in each layer. Their workspace and launch overheads are included in the reported latency.

\paragraph{Latency measurement.}
The five input-length/batch-size pairs are 256/1, 512/1, 1024/1, 1024/2, and 1024/4. Each request generates exactly 32 tokens with temperature 0 and EOS ignored. A local HTTP/SSE client measures TTFT and TPOT using actual emitted token IDs. Each configuration is tested in two rounds, with one warmup request followed by three timed requests per model in each round. The second round reverses model startup order; the reported values are medians pooled across both rounds.

\section{Properties of the two stages}
\label{app:proofs}

\paragraph{Exact solution of the capacity proxy.}
For a fixed width $k_e$ and positive $\rho_e$, a channel set omitting a higher-scoring channel in favor of a lower-scoring one cannot maximize Eq.~\eqref{eq:cape_allocation}. Hence an optimum exists in which each expert retains an anchor-score prefix. The lower-bound prefixes are mandatory. After retaining them, each eligible additional position has marginal value $\rho_e A_{e,(j)}$. Choosing the globally largest remaining marginals until the exact layer budget is reached maximizes the objective; upper-bound positions are excluded from the candidate pool. A zero multiplier gives equal marginal values for that expert, resolved by a fixed tie-break.

\paragraph{RAMP surrogate and exchanges.}
For a domain and expert, $Z^\top Z/N$ and $\widetilde W^\top\widetilde W$ are positive semidefinite. Their Hadamard product $G^{(d,e)}$ is positive semidefinite by the Schur product theorem, as is the nonnegative weighted sum across domains. Normalizing the Gram matrix on positive-diagonal channels preserves positive semidefiniteness. Zero-diagonal channels have zero off-diagonal entries and receive unit diagonal in $R$, so $R\succeq0$. For $0\le\beta\le1$, $(1-\beta)I+\beta R\succeq0$ and therefore $H=D_A[(1-\beta)I+\beta R]D_A\succeq0$.

For $p'=p+\mathbf e_i-\mathbf e_j$, expanding the quadratic form gives the exact objective change
\begin{equation}
\begin{aligned}
\Delta_{i\rightarrow j}={}&2(Hp)_i-2(Hp)_j\\
&+H_{ii}+H_{jj}-2H_{ij}.
\end{aligned}
\end{equation}
Accepting only $\Delta_{i\rightarrow j}<-\tau$ strictly lowers the surrogate. The simultaneous one-out and one-in update leaves the expert width unchanged. These statements concern the defined integer and reconstruction objectives, not model accuracy.

\makeatletter
\newcommand{\AppendixResultTable}[3]{%
\begingroup
\def\@captype{table}%
\captionsetup{hypcap=false}%
\centering
\small
\setlength{\tabcolsep}{2.5pt}%
\renewcommand{\arraystretch}{1.15}%
\begin{tabular*}{\textwidth}{@{\extracolsep{\fill}}l*{11}{r}@{}}
#1
\end{tabular*}\par
\caption{#2}
\label{#3}
\endgroup
}
\makeatother

\twocolumn[{%
\begin{minipage}{\textwidth}

\section{Full task results}
\label{app:fullresults}

Tables~\ref{tab:full1}--\ref{tab:full6} report the ten task accuracies and their unweighted mean for each model and pruning rate. WikiText-2 perplexity is used only for the Qwen3 example in Figure~\ref{fig:motivation} and the corresponding discussion in the main text.

\par\medskip

\AppendixResultTable{%
\toprule
Method & PIQA & ARC-E & ARC-C & GSM8K & Hella & Wino & RACE & GPQA-D & MATH & HE & Mean\\
\midrule
Dense & 91.29 & 94.40 & 93.00 & 96.29 & 87.12 & 75.30 & 91.61 & 56.57 & 80.60 & 95.73 & 86.19\\
CAMERA & 90.75 & 94.02 & 93.17 & 93.48 & 87.39 & 73.64 & 91.26 & 34.34 & 28.20 & 0.00 & 68.63\\
MoE-Slimming & 90.64 & 93.43 & 91.89 & 96.06 & 86.91 & 72.38 & 89.85 & 35.35 & 79.00 & 48.78 & 78.43\\
AIMER & 85.42 & 92.85 & 89.93 & 94.84 & 84.05 & 72.30 & 89.57 & 30.81 & 77.00 & 91.46 & 80.82\\
REAP & 89.01 & 93.90 & 92.15 & 96.06 & 86.27 & 71.74 & 89.05 & 43.43 & 78.20 & 53.05 & 79.29\\
MAN & 86.78 & 94.02 & 92.66 & 95.07 & 81.74 & 68.51 & 88.38 & 39.39 & 78.40 & 93.90 & 81.89\\
MSAN & 87.65 & 93.52 & 91.47 & 95.22 & 81.65 & 68.03 & 88.26 & 40.40 & 79.40 & 73.78 & 79.94\\
\addlinespace[2pt]
CAPE & 90.53 & 94.40 & 92.32 & 96.44 & 86.67 & 74.35 & 90.78 & 32.83 & 80.80 & 89.02 & 82.81\\
UC-RAMP & 90.70 & 94.40 & 92.49 & 96.59 & 86.76 & 73.80 & 90.90 & 33.84 & 79.00 & 89.63 & 82.81\\
\bottomrule
}{Task accuracy (\%) for Qwen3-30B-A3B-Instruct-2507; 25\%. MATH denotes MATH500 and HE denotes HumanEval pass@1. The 25\% or 50\% rate applies to routed-expert intermediate channels.}{tab:full1}

\par\bigskip

\AppendixResultTable{%
\toprule
Method & PIQA & ARC-E & ARC-C & GSM8K & Hella & Wino & RACE & GPQA-D & MATH & HE & Mean\\
\midrule
Dense & 91.29 & 94.40 & 93.00 & 96.29 & 87.12 & 75.30 & 91.61 & 56.57 & 80.60 & 95.73 & 86.19\\
CAMERA & 79.76 & 91.62 & 85.92 & 74.00 & 71.65 & 57.38 & 86.79 & 27.27 & 10.20 & 0.00 & 58.46\\
MoE-Slimming & 84.93 & 86.74 & 80.80 & 96.13 & 79.07 & 66.46 & 83.77 & 15.66 & 73.40 & 3.05 & 67.00\\
AIMER & 74.92 & 81.48 & 75.34 & 71.49 & 72.50 & 62.90 & 86.24 & 14.65 & 49.00 & 56.10 & 64.46\\
REAP & 79.87 & 91.46 & 87.12 & 93.63 & 72.75 & 59.67 & 83.82 & 21.72 & 64.40 & 8.54 & 66.30\\
MAN & 65.13 & 85.65 & 79.78 & 90.60 & 46.44 & 55.56 & 72.50 & 34.34 & 81.00 & 37.20 & 64.82\\
MSAN & 68.23 & 88.01 & 81.48 & 92.19 & 49.56 & 55.25 & 66.56 & 34.85 & 81.40 & 54.88 & 67.24\\
\addlinespace[2pt]
CAPE & 82.43 & 92.13 & 88.48 & 95.45 & 81.71 & 68.35 & 89.12 & 21.72 & 77.00 & 55.49 & 75.19\\
UC-RAMP & 82.43 & 92.09 & 88.82 & 95.53 & 81.70 & 68.51 & 88.91 & 19.19 & 78.00 & 59.15 & 75.43\\
\bottomrule
}{Task accuracy (\%) for Qwen3-30B-A3B-Instruct-2507; 50\%. MATH denotes MATH500 and HE denotes HumanEval pass@1. The 25\% or 50\% rate applies to routed-expert intermediate channels.}{tab:full2}

\par\vspace{6pt}
\end{minipage}%
}]

\clearpage

\twocolumn[{%
\begin{minipage}{\textwidth}

\AppendixResultTable{%
\toprule
Method & PIQA & ARC-E & ARC-C & GSM8K & Hella & Wino & RACE & GPQA-D & MATH & HE & Mean\\
\midrule
Dense & 93.42 & 94.95 & 95.39 & 96.66 & 79.43 & 85.40 & 92.56 & 72.73 & 94.00 & 96.95 & 90.15\\
CAMERA & 90.70 & 94.91 & 95.31 & 94.24 & 78.38 & 81.77 & 91.87 & 43.43 & 20.60 & 0.00 & 69.12\\
MoE-Slimming & 87.32 & 91.25 & 89.76 & 94.54 & 73.32 & 73.95 & 88.36 & 39.39 & 75.20 & 85.37 & 79.85\\
AIMER & 83.24 & 84.13 & 82.59 & 62.47 & 73.64 & 76.95 & 91.57 & 22.73 & 36.20 & 9.15 & 62.27\\
REAP & 88.47 & 94.23 & 92.06 & 95.91 & 81.51 & 71.43 & 90.78 & 60.10 & 89.40 & 33.54 & 79.74\\
MAN & 65.02 & 87.46 & 86.77 & 43.21 & 55.84 & 55.88 & 75.36 & 54.04 & 85.20 & 90.85 & 69.96\\
MSAN & 87.60 & 93.64 & 91.72 & 95.91 & 76.08 & 68.43 & 90.88 & 57.07 & 81.60 & 96.95 & 83.99\\
\addlinespace[2pt]
CAPE & 84.55 & 94.57 & 94.37 & 96.59 & 73.17 & 78.06 & 91.51 & 56.57 & 91.80 & 87.80 & 84.90\\
UC-RAMP & 84.93 & 94.32 & 94.45 & 96.59 & 74.18 & 77.58 & 91.41 & 56.57 & 91.20 & 88.41 & 84.96\\
\bottomrule
}{Task accuracy (\%) for Gemma4-26B-A4B-it; 25\%. MATH denotes MATH500 and HE denotes HumanEval pass@1. The 25\% or 50\% rate applies to routed-expert intermediate channels.}{tab:full3}

\par\bigskip

\AppendixResultTable{%
\toprule
Method & PIQA & ARC-E & ARC-C & GSM8K & Hella & Wino & RACE & GPQA-D & MATH & HE & Mean\\
\midrule
Dense & 93.42 & 94.95 & 95.39 & 96.66 & 79.43 & 85.40 & 92.56 & 72.73 & 94.00 & 96.95 & 90.15\\
CAMERA & 69.80 & 89.73 & 84.56 & 63.38 & 70.30 & 61.17 & 84.48 & 33.33 & 4.40 & 0.00 & 56.12\\
MoE-Slimming & 77.86 & 75.72 & 67.66 & 90.60 & 64.60 & 61.01 & 82.47 & 31.82 & 58.20 & 29.27 & 63.92\\
AIMER & 64.25 & 57.66 & 52.22 & 15.31 & 56.52 & 62.04 & 86.71 & 19.19 & 4.40 & 0.00 & 41.83\\
REAP & 13.11 & 25.72 & 23.89 & 86.13 & 14.63 & 11.13 & 74.34 & 17.68 & 12.60 & 0.00 & 27.92\\
MAN & 55.17 & 70.41 & 61.69 & 74.68 & 33.26 & 26.20 & 75.74 & 24.75 & 58.00 & 12.20 & 49.21\\
MSAN & 49.78 & 53.87 & 50.77 & 84.76 & 30.29 & 33.86 & 83.91 & 24.24 & 61.40 & 7.93 & 48.08\\
\addlinespace[2pt]
CAPE & 73.45 & 89.73 & 85.67 & 94.84 & 62.70 & 63.93 & 86.32 & 31.31 & 73.40 & 14.63 & 67.60\\
UC-RAMP & 72.69 & 90.53 & 87.20 & 95.45 & 63.50 & 66.54 & 86.81 & 28.28 & 73.80 & 14.02 & 67.88\\
\bottomrule
}{Task accuracy (\%) for Gemma4-26B-A4B-it; 50\%. MATH denotes MATH500 and HE denotes HumanEval pass@1. The 25\% or 50\% rate applies to routed-expert intermediate channels.}{tab:full4}

\par\vspace{6pt}
\end{minipage}%
}]

\clearpage

\twocolumn[{%
\begin{minipage}{\textwidth}

\AppendixResultTable{%
\toprule
Method & PIQA & ARC-E & ARC-C & GSM8K & Hella & Wino & RACE & GPQA-D & MATH & HE & Mean\\
\midrule
Dense & 94.34 & 94.15 & 93.86 & 86.50 & 90.44 & 91.32 & 89.60 & 20.20 & 24.80 & 84.15 & 76.94\\
CAMERA & 95.16 & 94.78 & 95.48 & 87.19 & 91.17 & 92.82 & 92.70 & 16.67 & 31.20 & 1.83 & 69.90\\
MoE-Slimming & 89.39 & 90.03 & 91.64 & 41.93 & 82.24 & 83.66 & 79.43 & 11.62 & 15.60 & 61.59 & 64.71\\
AIMER & 91.95 & 90.78 & 90.61 & 73.16 & 82.70 & 88.24 & 84.99 & 11.62 & 29.20 & 66.46 & 70.97\\
REAP & 94.67 & 92.72 & 94.20 & 75.44 & 90.76 & 91.08 & 89.68 & 12.12 & 27.40 & 25.61 & 69.37\\
MAN & 94.67 & 92.72 & 93.94 & 73.77 & 90.55 & 88.63 & 90.82 & 15.66 & 24.40 & 77.44 & 74.26\\
MSAN & 94.34 & 93.73 & 94.88 & 75.97 & 91.03 & 90.61 & 91.95 & 18.18 & 27.80 & 77.44 & 75.59\\
\addlinespace[2pt]
CAPE & 92.00 & 92.38 & 90.78 & 77.79 & 80.49 & 86.50 & 83.68 & 3.03 & 28.80 & 72.56 & 70.80\\
UC-RAMP & 91.13 & 91.50 & 90.78 & 77.18 & 82.20 & 85.64 & 83.04 & 5.56 & 29.40 & 35.37 & 67.18\\
\bottomrule
}{Task accuracy (\%) for Qwen3.6-35B-A3B; 25\%. MATH denotes MATH500 and HE denotes HumanEval pass@1. The 25\% or 50\% rate applies to routed-expert intermediate channels.}{tab:full5}

\par\bigskip

\AppendixResultTable{%
\toprule
Method & PIQA & ARC-E & ARC-C & GSM8K & Hella & Wino & RACE & GPQA-D & MATH & HE & Mean\\
\midrule
Dense & 94.34 & 94.15 & 93.86 & 86.50 & 90.44 & 91.32 & 89.60 & 20.20 & 24.80 & 84.15 & 76.94\\
CAMERA & 85.26 & 88.85 & 89.93 & 33.74 & 59.62 & 81.93 & 89.58 & 20.20 & 12.20 & 0.00 & 56.13\\
MoE-Slimming & 71.38 & 86.95 & 82.17 & 20.24 & 66.71 & 67.01 & 73.19 & 6.57 & 11.00 & 10.98 & 49.62\\
AIMER & 65.18 & 67.30 & 63.40 & 13.80 & 49.22 & 79.79 & 57.70 & 25.76 & 43.00 & 23.78 & 48.89\\
REAP & 90.21 & 88.05 & 86.69 & 89.84 & 78.71 & 88.71 & 86.90 & 18.18 & 25.60 & 1.22 & 65.41\\
MAN & 90.81 & 92.21 & 91.98 & 90.30 & 80.38 & 87.37 & 88.55 & 8.59 & 20.20 & 30.49 & 68.09\\
MSAN & 89.45 & 92.47 & 91.89 & 75.44 & 81.67 & 87.21 & 90.01 & 12.12 & 20.40 & 29.88 & 67.05\\
\addlinespace[2pt]
CAPE & 76.44 & 86.99 & 88.05 & 93.10 & 75.91 & 85.56 & 89.18 & 8.08 & 52.20 & 71.34 & 72.69\\
UC-RAMP & 78.78 & 88.09 & 88.91 & 93.40 & 76.89 & 85.40 & 89.74 & 12.63 & 52.80 & 75.61 & 74.23\\
\bottomrule
}{Task accuracy (\%) for Qwen3.6-35B-A3B; 50\%. MATH denotes MATH500 and HE denotes HumanEval pass@1. The 25\% or 50\% rate applies to routed-expert intermediate channels.}{tab:full6}

\par\vspace{6pt}
\end{minipage}%
}]

\section{Additional component comparisons}
\label{app:ablations}

Table~\ref{tab:ablation_extra} reports the four remaining model-budget settings, complementing the two 50\% component ablations in Table~\ref{tab:ablation}. Every CAPE--UC-RAMP pair uses the same expert widths.

\begin{table}[!htbp]
\centering
\small
\begin{tabular}{lrrr}
\toprule
Model / ratio & CAPE & UC-RAMP & $\Delta$ RAMP\\
\midrule
Qwen3, 25\% & 82.81 & 82.81 & 0.00\\
Gemma-4, 25\% & 84.90 & 84.96 & +0.06\\
Qwen3.6, 25\% & 70.80 & 67.18 & -3.62\\
Qwen3.6, 50\% & 72.69 & 74.23 & +1.54\\
\bottomrule
\end{tabular}
\caption{Additional fixed-width refinement comparisons in ten-task mean accuracy (\%). $\Delta$ RAMP is the difference between the displayed UC-RAMP and CAPE means.}
\label{tab:ablation_extra}
\end{table}

On Qwen3.6 at 50\%, refinement increases GPQA-Diamond accuracy from 8.08\% to 12.63\% and HumanEval pass@1 from 71.34\% to 75.61\%, while MATH500 changes from 52.2\% to 52.8\%. At 25\%, HumanEval changes from 72.56\% to 35.37\% and MATH500 from 28.8\% to 29.4\%. Appendix~\ref{app:fullresults} provides the complete task scores for these configurations.

\end{document}